\documentclass[conference]{IEEEtran}
\usepackage{cite}
\usepackage{amsmath,amssymb,amsfonts}
\usepackage{algorithmic}
\usepackage{graphicx}
\usepackage{textcomp}
\usepackage{xcolor}

\usepackage{graphics} % for pdf, bitmapped graphics files
\usepackage{epsfig} % for postscript graphics files
\usepackage{mathptmx} % assumes new font selection scheme installed
\usepackage{times} % assumes new font selection scheme installed
\usepackage{xspace}
\usepackage[pagebackref,breaklinks]{hyperref}

\def\BibTeX{{\rm B\kern-.05em{\sc i\kern-.025em b}\kern-.08em
    T\kern-.1667em\lower.7ex\hbox{E}\kern-.125emX}}

\newcommand{\our}{ECAM\xspace}
\newcommand{\ournospace}{ECAM}
\newcommand{\mapnce}{MapNCE\xspace}
\newcommand{\mapncenospace}{MapNCE}

\title{\LARGE \our: A Contrastive Learning Approach to Avoid \\Environmental Collision in Trajectory Forecasting}

\author{
    \IEEEauthorblockN{Giacomo Rosin$^{1}$, Muhammad Rameez Ur Rahman$^{1}$, Sebastiano Vascon$^{1,2}$}
    \IEEEauthorblockA{
        \textit{Department of Environmental Sciences, Informatics and Statistics, Ca' Foscari University of Venice, Italy$^{1}$}}
    \IEEEauthorblockA{
        \textit{European Centre for Living Technology$^{2}$}
    \\
        \{giacomo.rosin, muhammad.rahman, sebastiano.vascon\}@unive.it
    }
}

\begin{document}

\maketitle
\thispagestyle{empty}
\pagestyle{empty}

%%%%%%%%%%%%%%%%%%%%%%%%%%%%%%%%%%%%%%%%%%%%%%%%%%%%%%%%%%%%%%%%%%%%%%%%%%%%%%%%
\begin{abstract}
Human trajectory forecasting is crucial in applications such as autonomous driving, robotics and surveillance. Accurate forecasting requires models to consider various factors, including social interactions, multi-modal predictions, pedestrian intention and environmental context. While existing methods account for these factors, they often overlook the impact of the environment, which leads to collisions with obstacles. This paper introduces \our (Environmental Collision Avoidance Module), a contrastive learning-based module to enhance collision avoidance ability with the environment. The proposed module can be integrated into existing trajectory forecasting models, improving their ability to generate collision-free predictions. We evaluate our method on the ETH/UCY dataset and quantitatively and qualitatively demonstrate its collision avoidance capabilities. Our experiments show that state-of-the-art methods significantly reduce (-40/50\%) the collision rate when integrated with the proposed module. The code is available at \url{https://github.com/CVML-CFU/ECAM}.
\end{abstract}

%%%%%%%%%%%%%%%%%%%%%%%%%%%%%%%%%%%%%%%%%%%%%%%%%%%%%%%%%%%%%%%%%%%%%%%%%%%%%%%%
\section{INTRODUCTION}
Human trajectory forecasting aims to predict the future movements of individuals based on their past trajectories and interactions with the environment. Forecasting pedestrian trajectories is crucial for applications such as self-driving cars \cite{gulzar2021survey}, social robotics \cite{socialrobot}, and surveillance \cite{ano_pred}. For example, self-driving cars use future pedestrian movements to prevent accidents and ensure the safety of passengers and nearby individuals.

Forecasting pedestrian trajectories is inherently complex due to the wide range of human behaviors, social interactions, individual intent, and environmental context.  A good forecasting method should consider these factors to forecast close to natural trajectories. Several proposed methods \cite{alahi2016social, sun2020recursive, zhang2019sr, transformer, liu2021social} model social interactions to prevent collisions and forecast trajectories for people walking in groups. On the other hand, several goal-oriented methods \cite{mangalam2020not, mangalam2021goals, wang2022stepwise, chen2023goal} are proposed to exploit individuals intent. Furthermore, naturally, humans can have many plausible futures, so a good prediction model must consider all these possible paths and generate diverse predictions \cite{gupta2018social, dendorfer2021mg, transformer}. While all these methods exploit social interactions between pedestrians, individual intent, and multi-modal predictions, their consideration of scene context is often limited.

\begin{figure}[t!]
  \centering
  \framebox{\includegraphics[scale=0.32]{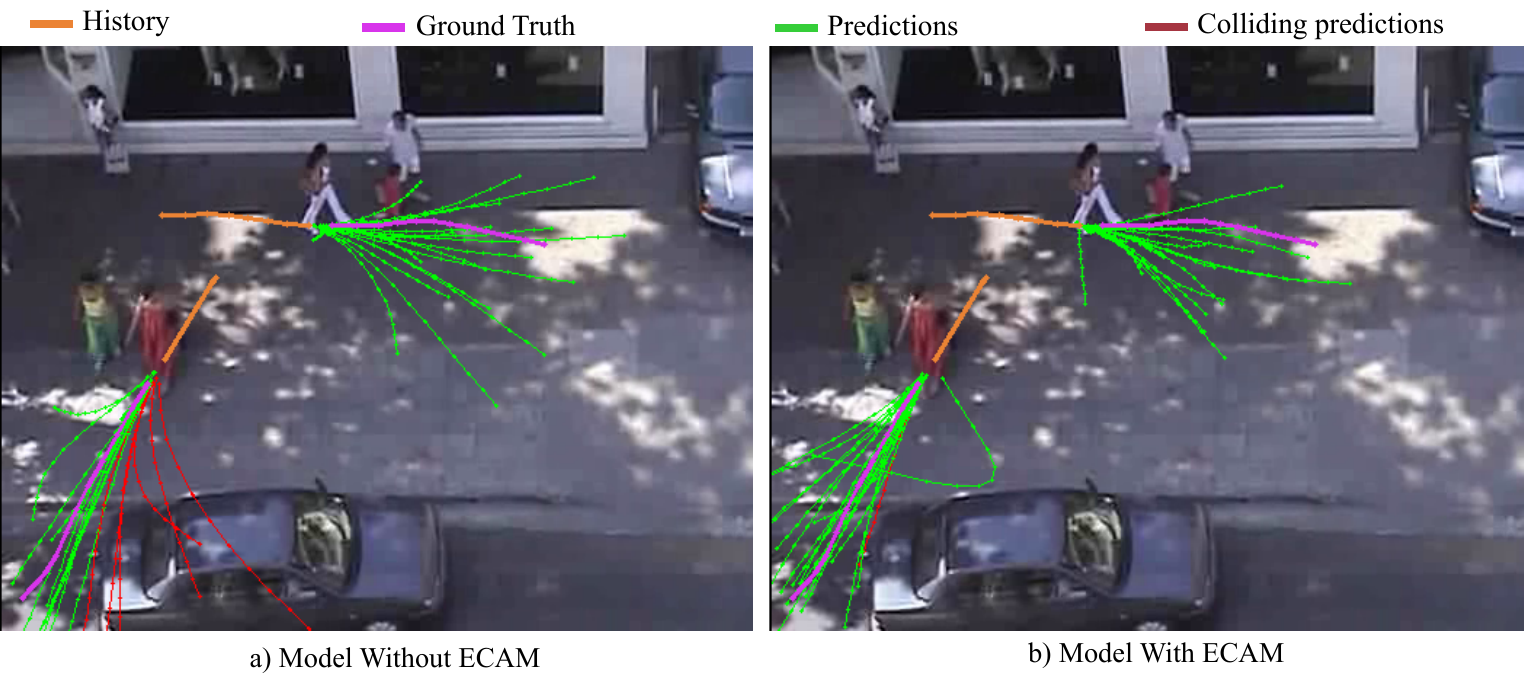}}
  \caption{Trajectory forecasting on ZARA1 dataset a) Model \cite{bae2024singulartrajectory} without \our, b) Model with \our. A model integrated with \our shows no or minimal number of collisions with the environment.}
  \label{teaser}
\end{figure}

The surrounding context plays a crucial role in how people safely navigate in an environment to avoid obstacles. Several pedestrian trajectory forecasting methods use scene context \cite{kosaraju2019social, bae2024singulartrajectory, yue2022human, yuan2021agentformer, rempe2023trace, lee2022muse, liu2024trajdiffuse} to model interaction with the environment. However, existing approaches either fail to address the prevention of collisions, as shown in Fig.\ref{teaser}a, or rely on computationally expensive or hand-crafted operations. Hence, we focus on explicitly learning environmental collision avoidance, in a more efficient and scalable way.

In this paper, we introduce \our (Environmental Collision Avoidance Module), which consists of \mapnce, a contrastive learning-based module, and a supplementary Environmental Collision loss (EnvColLoss). The \mapnce module (Map Noise-Contrastive Estimation), inspired by Social-NCE \cite{liu2021social} (which focuses only on social interactions), encourages the model to encode information about the surrounding environment in the hidden representations. It uses the environment map to automatically generate negative samples (areas near obstacles) for training the model in a self-supervised contrastive way. The key idea is to train the model in such a way that it can distinguish between a pedestrian's actual future trajectory and trajectories that lead to collisions with environmental obstacles. The usefulness of this module comes from the fact that it can exploit these additional samples instead of just using the available positive samples (points belonging to the true trajectory) contained in the dataset.
Beyond the contrastive loss from the \mapnce module, \our incorporates a supplementary Environmental Collision loss to ensure the model effectively utilizes learned representations for collision avoidance across all predicted trajectory samples. This loss penalizes the model when any of the sampled trajectories collide with the obstacles in the scene.

The proposed module can be integrated into any existing trajectory forecasting model to improve its collision avoidance ability with the environment without any overhead at inference time since it is only used during training. We evaluate our method on the ETH/UCY \cite{pellegrini2009you, lerner2007crowds} dataset, which provides bird's-eye view data of pedestrian interactions, and demonstrate that the proposed module significantly reduces the collision rate when integrated with state-of-the-art methods.

The main contributions of this paper are as follows:
\begin{itemize}
\item We propose a contrastive learning approach paired with an additional Environmental Collision loss that enhances the model's spatial reasoning capabilities to avoid collisions with environmental constraints.
\item \our is only used at training time and can be integrated into any map-based trajectory forecasting methods without introducing any computational overhead.
\item We integrate \our into state-of-the-art methods and quantitatively and qualitatively show that it significantly improves their collision avoidance ability.
\end{itemize}

\section{Related Work}
\subsection{Trajectory forecasting}
Early frameworks for trajectory prediction were based on hand-crafted rules, such as attraction to goals, repulsion from obstacles and other agents \cite{helbing1995social, van2008reciprocal}. They perform reasonably on interaction modelling but offer poor generalization. Since trajectories are sequences, recurrent neural networks (RNNs) are a
natural choice, and pioneering works \cite{alahi2016social, lee2017desire,
gupta2018social, sadeghian2019sophie, kosaraju2019social,
salzmann2020trajectron++} used them as encoders and decoders, predicting the future in an autoregressive way. Similarly, with the success of Transformers \cite{vaswani2017attention} in modelling sequential data, 
several works have proposed to use them for trajectory prediction \cite{franco2023under, liu2021multimodal, yuan2021agentformer}. Notably, \cite{yuan2021agentformer} proposes a model that simultaneously integrates temporal and social information using a novel agent-aware attention. With the recent success of diffusion models \cite{ho2020denoising, nichol2021improved} in image generation, \cite{gu2022stochastic} proposed to use them for trajectory prediction. These models predict the whole future trajectory at once, improving accuracy by capturing long-term dependencies
Distinctively, \cite{bae2024singulartrajectory} proposes a diffusion-based framework that can handle different types of trajectory forecasting tasks using a shared motion embedding space.

\subsection{Scene-Aware Trajectory forecasting}
The surrounding context plays a crucial role in how people move
in an environment, as it places hard constraints on their movements.
To condition the trajectory prediction on the scene context, previous works \cite{lee2017desire, xue2018ss, sadeghian2019sophie,
kosaraju2019social, li2019conditional, dendorfer2021mg, yuan2021agentformer, rempe2023trace} used a Convolutional Neural Network (CNN) to encode the full image or patches of the image into a vector for conditioning the trajectory generation.
Notably, \cite{sadeghian2019sophie, kosaraju2019social, li2019conditional}
use the attention mechanism to focus on the relevant parts of the scene. However, compressing the scene image into a one-dimensional space results in losing spatial information and the alignment between the encoded trajectories and the image.
Y-Net \cite{mangalam2021goals} and later \cite{lee2022muse, liu2024trajdiffuse} solved the alignment issue by representing trajectories on the same two-dimensional space as the scene.
Another approach is proposed in \cite{bae2024singulartrajectory, liu2024trajdiffuse}, which represents the scene as a vector field that guides the trajectories toward the nearest obstacle-free zones.
While these methods consider scene constraints, they either rely on computationally intensive operations \cite{shukla2022from} or employ non-learned, hand-crafted trajectory adjustments that may limit their scalability.
Hence, this paper proposes a learnable approach based on contrastive learning combined with a novel
environmental collision loss that explicitly accounts for the interaction between pedestrians and surrounding environment to generate collision-free trajectories, without the need for additional computational overhead at inference time.

\section{Methodology}
\begin{figure*}[th!]
  \centering
  %\framebox{\parbox{\textwidth}{Add here the pipeline of your module\\!\\!\\!\\!\\!\\!\\!\\!\\}}
  \includegraphics[width=0.95\textwidth]{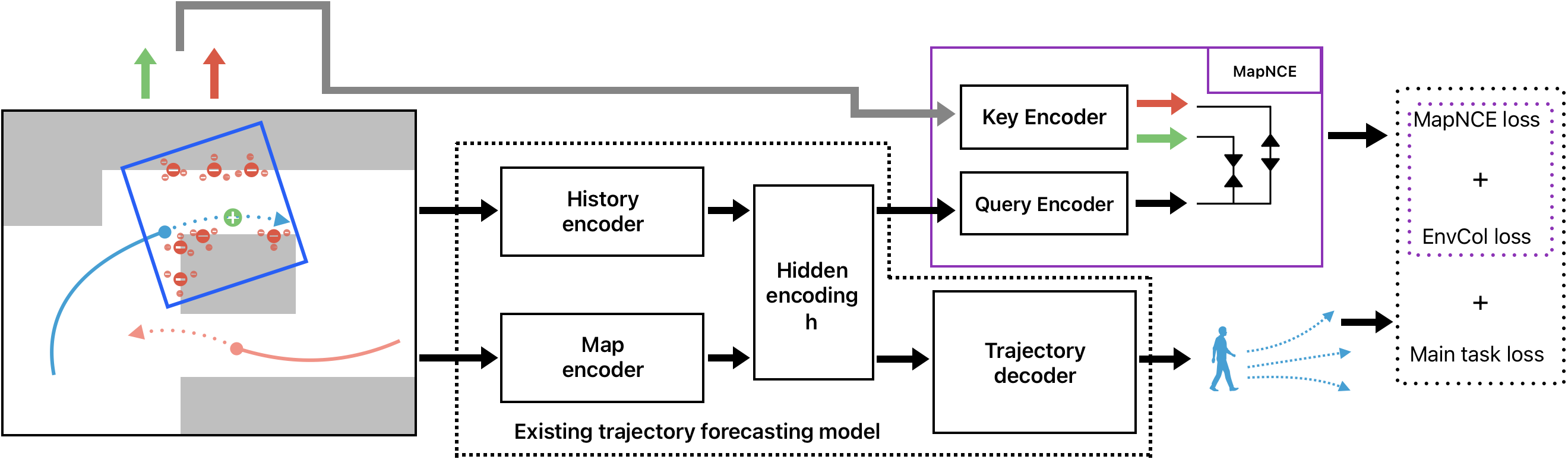}
  \caption{\our is a training time module that can be added on top of an existing trajectory forecasting model, given that it computes an embedding of the scene map (and of the history of each pedestrian); the hidden representation is then forwarded to the query encoder of the \mapnce module, and at the same time a positive sample belonging to the future trajectory, together with some negative samples close to obstacles are forwarded to the key encoder. The goal of the \mapnce module is to be able to tell apart positive and negative samples given the query. Finally, the predicted trajectories that collide with obstacles are penalized through the Environmental Collision loss (EnvColLoss). The figure also highlights the negative sampling extraction around the pedestrian.}
  \label{fig:module}
\end{figure*}

In this section, we first explain the problem formulation. Then, we describe the building blocks of the proposed Environmental Collision Avoidance Module (\ournospace), shown in Figure \ref{fig:module}, particularly the \mapnce module, its loss function, and the Environmental Collision loss.
\subsection{Problem Formulation}
The input to the model consists of all pedestrians' past positions, represented as a sequence of 2D coordinates. The input can be denoted as $\mathbf{X} = \{X_1, X_2, \ldots, X_N\}$, where $N$ is the total number of pedestrians in the scene and 
$X_i = [x_{i,t} \in \mathbb{R}^2 \mid t = 1, 2, \ldots, T_{\text{obs}}]$ is the sequence of 2D coordinates of pedestrian $i$ with an observation window of size $T_{\text{obs}}$. The model can also take additional information, such as the images $\mathbf{I} = [I_{t} \mid t = 1, 2, \ldots, T_{\text{obs}}]$ of the scene at each time step. In our case, we assume we have access to a segmentation mask of the scene that encodes the walkable area and the obstacles in the scene. Moreover, we use only the last image of the sequence, $I_{T_{\text{obs}}}$, since it contains the most up-to-date information to base the prediction on.

The task of the model is to forecast the corresponding sequence of future positions $\mathbf{Y} = \{Y_1, Y_2, \ldots, Y_N\}$ of all the pedestrians in the scene, where $Y_i = [y_{i,t} \in \mathbb{R}^2 \mid t = T_{\text{obs}} + 1, T_{\text{obs}} + 2, \ldots, T_{\text{obs}} + T_{\text{pred}}]$ is the future sequence of 2D coordinates of pedestrian $i$ for a prediction window of size $T_{\text{pred}}$. The model output is denoted as $\mathbf{\hat{Y}} = \{\hat{Y}_1, \hat{Y}_2, \ldots, \hat{Y}_N\}$. However, given the multi-modal nature of the human trajectories, the model is required to generate $K$ future trajectories for each pedestrian, in which case its output is a set of $K$ sets of $N$ future trajectories.

\subsection{\mapnce Module}
\label{subsec:map_nce_module}
To learn a robust representation that can generalize to new environments, the model must be able to distinguish between a pedestrian's actual future trajectory and zones around the obstacles in the scene. The \mapnce module forces the learned map and trajectory representations to contain useful information about the spatial structure of the scene in order to avoid collisions with the obstacles.

To be able to distinguish between future valid positions (positive samples)
and collision zones (negative samples) of a given pedestrian
the \mapnce module needs to get information from the underlying model about
what the past trajectory of the pedestrian looks like and how the map is composed.
So, when adding the \mapnce module to a model, the model must be both conditioned
on the past trajectory of the pedestrian and on the map of the scene.

The \mapnce module is composed of two submodules: a query encoder and a key encoder. The query encoder processes the model's hidden representation for each pedestrian through a linear projection head, generating query vectors. The key encoder maps both positive and negative samples through an MLP to produce the corresponding key vectors. These vectors are embedded in a shared space, where similarity is measured via dot products.

The procedure to obtain the positive and negative samples for pedestrian $i$
is as follows.
Given a future timestep $t$, the positive sample is the true future
position of the pedestrian $i$ at timestep $t$, with a small Gaussian noise
to prevent overfitting, denoted by:
\begin{equation}
  \text{pos\_sample}_i = y_i^t + \epsilon
  \label{eq:map_positive_sample}
\end{equation}
where $\epsilon \sim \mathcal{N}(0, c_{\epsilon}I)$, and $c_{\epsilon}$
is a small constant, set to 0.05 meters in our experiments.
The negative samples are obtained starting from the contours of the obstacles (set of points) in the environment map, as these represent the critical boundaries between navigable and non-navigable areas, making points near them informative negative examples for collision avoidance.
The set of contour points for pedestrian $i$ is
$\{c_{i,1}, c_{i,2}, \ldots, c_{i,M_i}\}$
where $M_i$ is the number of contour points for pedestrian $i$.
For each pedestrian $i$, we sample a subset of contour points $Z = 10$,
denoted by $\text{neg\_sample\_seeds}_i$:
\begin{equation}
  \text{neg\_sample\_seeds}_i = \{c_{i,1}, c_{i,2}, \ldots, c_{i,Z}\}
  \label{eq:map_negative_samples}
\end{equation}
The $\text{neg\_sample\_seeds}_i$ are used to generate 8 negative samples
around each contour point $c_{i,z} \in \text{neg\_sample\_seeds}_i$
by adding a small displacement $\Delta_p = (\rho \cos \theta_p, \rho \sin \theta_p)$:
\begin{equation}
  \text{neg\_samples}_{i,z} = c_{i,z} + \Delta_p + \epsilon
  \label{eq:map_negative_sample}
\end{equation}
where $\rho$ is a small constant, set to 0.5 meters in our experiments,
representing a plausible proximity to an obstacle that should be avoided,
$\theta_p = \frac{\pi}{4}p$ is the angle of the displacement that depends on the 
index $p$ which ranges from 0 to 7, corresponding to the 8 displacement directions
around the contour point.
Again, a small Gaussian noise $\epsilon$ is added to the negative samples to
prevent overfitting.
Finally, the set of negative samples for pedestrian $i$ is the union of the
negative samples around all the $Z$ selected contour points in the scene:
\begin{equation}
  \text{neg\_samples}_i = \bigcup_{z \in \{1, 2, \ldots, Z\}
  } \text{neg\_samples}_{i,z}
  \label{eq:map_negative_samples_union}
\end{equation}

In order to perform the contrastive learning, the \mapnce module defines
a query encoder $\psi$ that embeds the hidden representation $h_i$ generated by the encoder of
the underlying model for each pedestrian $i$ in the scene into a query vector $q_i$:
\begin{equation}
  q_i = \psi(h_i) = \text{Linear}_{\text{query}}(h_i; W_{\text{query}})
  \label{eq:map_query_encoder}
\end{equation}
and a key encoder $\phi$ that embeds the positive and negative samples into a set
of key vectors $K_i = \{k_{i,0}, k_{i,1}, \ldots, k_{i,Z \times 8}\}$ 
for each individual $i$ in the scene, 
where $k_{i,0}$ is the key corresponding to the positive sample, and
$k_{i,z}$ are the keys of the negative samples with index 
$z \in \{1, 2, \ldots, Z \times 8\}$:
\begin{align}
    k_{i,0} &= \phi(\text{pos\_sample}_i) = \text{MLP}_{\text{key}}(\text{pos\_sample}_i; W_{\text{key}}) \\
    k_{i,z} &= \phi(\text{neg\_samples}_{i,z}) = \text{MLP}_{\text{key}}(\text{neg\_samples}_{i,z}; W_{\text{key}})
  \label{eq:map_key_encoder}
\end{align}
where $W_{\text{query}}$ and $W_{\text{key}}$ are the parameters of the
MLP models.

The goal of the \mapnce module is to maximize the similarity between
the query vector $q_i$ and the key vector $k_{i,0}$ corresponding to the
positive sample, while minimizing the similarity between the query vector
$q_i$ and the key vectors $k_{i,z}$ corresponding to the negative samples.
This is achieved by optimizing the \mapnce contrastive loss
detailed in section \ref{subsec:map_nce_losses}.

%%%%%%%%%%%%%%%%%%%%%%%%%%%%%%%%%%%%%%%%%%%%%%%%%%%%%%%%%%%%%%%%%%%%%%%%%%%%%%%%
\subsection{Loss}
\subsubsection{\mapnce Loss}
\label{subsec:map_nce_losses}
\mapnce loss is the auxiliary contrastive loss associated with the \mapnce module to 
guide the model in learning which zones of the map it should avoid 
generating prediction in. It encourages the model to encode information about the 
surrounding environment in the hidden representations,
by learning in a self-supervised way from contrastive examples (points near the obstacles that we would like the pedestrians to avoid) generated automatically by leveraging domain knowledge about obstacles. Without this loss, the model would learn only what is
correct, and would need to learn what to avoid by pure generalization, whereas \mapnce gives it a push in the right direction by supervising its training with additional 
useful signals.

The \mapnce loss function for a single pedestrian $i$, is given by:
\begin{equation}
  \mathcal{L}_\textbf{\mapnce}^i =
  - \log
  \frac
  {\exp(\psi(h_i) \cdot \phi(\text{pos\_sample}_{i}) / \tau)}
  {\sum \limits_{j=0}^J \exp (\psi(h_i) \cdot \phi(\text{samples}_{i,j}) / \tau)}
  \label{eq:individual_nce_loss}
\end{equation}
where $\psi$ is the query encoder, $\phi$ is the key encoder, $\tau$ is the temperature
parameter, and $J$ is the number of negative samples. Recall also that $h_i$ is the
hidden representation generated by the underlying model encoder, $\text{pos\_sample}_{i} = \text{samples}_{i, 0}$ is the positive sample, and $\text{samples}_{i,j>0}$ are the negative samples.
Finally, the average over all the pedestrians in the scene is taken.
This loss is basically the binary cross-entropy loss for the classification problem of distinguishing the positive key from the negative keys, given the query.

\subsubsection{Environmental Collision Loss}
\label{subsec:environment_collision_loss}
To encourage the model to learn obstacle avoidance, we introduce an environmental collision loss, which penalizes the model when the sampled trajectories collide with obstacles in the scene. Unlike variety loss \cite{gupta2018social}, commonly used as an optimization objective in multimodal trajectory forecasting \cite{yuan2021agentformer, bae2024singulartrajectory}, which backpropagates gradients only for the best trajectory, the environmental collision loss computes the mean squared error for all the colliding trajectories. As a result, the model learns obstacle avoidance across all samples.

For a pedestrian $i$, the set of colliding samples is defined as:
$C_i = \{k \in \{1, 2, \ldots, K\} \mid \text{collides}(\hat{Y}^k_i, I_i)\}$,
where $\text{collides}(\hat{Y}^k_i, I_i)$ is defined as:
\begin{equation}
  \text{collides}(\hat{Y}^k_i, I) =
  \begin{cases}
    \text{True} & \text{if } \exists\mbox{ } t \in T \text{ s.t. } I(g(\hat{y}^k_{i,t})) = 0 \\
    \text{False} & \text{otherwise}
  \end{cases}
  \label{eq:metric_collides}
\end{equation}
where $g(\hat{y}^k_{i,t})$ is the function that maps the trajectory point
$\hat{y}^k_{i,t}$ to the corresponding pixel coordinates,
$I(g(\hat{y}^k_{i,t}))$ is the value of the pixel in the image $I$ at
the trajectory position $\hat{y}^k_{i,t}$, and
$T = \{T_{\text{obs}} + 1, \ldots, T_{\text{obs}} + T_{\text{pred}}\}$ 
is the set of future time steps.
The environmental collision loss for a scene with $N$ pedestrians is defined as:
\begin{equation}
  \mathcal{L}_{\text{Env-Col}} = \frac{1}{N}
  \sum_{i=1}^{N}
  \frac{1}{|C_i|}
  \sum_{k \in C_i}
  \left\| \hat{Y}^k_i - Y_i \right\|^2_2
  \label{eq:environment_collision_loss}
\end{equation}
If $|C_i| = 0$, that is, pedestrian $i$ doesn't collide with any obstacle, then its own
loss is considered to be $0$.

\section{Experiments}
We validate our approach by comparing three trajectory forecasting models,
namely SingularTrajectory (ST) \cite{bae2024singulartrajectory} (diffusion-based),
EigenTrajectory-AgentFormer (E-AF) \cite{yuan2021agentformer, bae2023eigentrajectory} (transformer-based),
and EigenTrajectory-SGCN (E-SGCN) \cite{shi2021sgcn, bae2023eigentrajectory} (graph convolutional network-based),
with their enhanced versions incorporating our proposed approach (see "+\ournospace" in the results).
We choose these models because they are built on diverse modern architectures (graphs, transformers and diffusion)
and %have shown to be high-performing 
are state-of-the-art in the trajectory forecasting task.
This enables us to assess the generality of our approach.

\subsection{Dataset}
We use ETH/UCY dataset \cite{pellegrini2009you, lerner2007crowds}, a standard benchmark to evaluate human trajectory prediction. It comprises five bird's-eye view scenes: ETH (eth and hotel) and UCY (univ, zara1, zara2), which include over $1500$ pedestrian trajectories with many scenarios. To ensure fair comparisons, we adopt the leave-one-out approach \cite{alahi2016social} that is standard in the literature, where the models are trained on four scenes and tested on the fifth, with the results averaged across all scenes.
\subsection{Evaluation Metrics}
We evaluate the models on the ETH/UCY dataset using the standard
Average and Final Displacement Error ($\text{ADE}_\text{min}$/$\text{FDE}_\text{min}$)
\cite{gupta2018social, yuan2021agentformer}
which give a measure of the distance between predicted and ground truth trajectories.
We then use a recently proposed Environment Collision-Free Likelihood (ECFL) metric \cite{sohn2021a2x}, that evaluates the ability of the model to avoid collisions with the obstacles in the scene.
\subsubsection{Average and Final Displacement Error (ADE)}
\label{subsec:ade_fde_metric}
The Average Displacement Error (ADE) measures the average L2
distance between predicted and ground truth trajectory at
each future time step (lower is better). The ADE for a single pedestrian $i$ is:
\begin{equation}
  \text{ADE}_i =
  \frac{1}{T_{\text{pred}}}
  \sum_{t=T_{\text{obs}+1}}^{T_{\text{obs}}+T_{\text{pred}}}
  \left\| \hat{y}_{i,t} - y_{i,t} \right\|_2
  \label{eq:ade_single}
\end{equation}
where $T_{\text{obs}}$ is the time of the last observed step, $T_{\text{pred}}$
is the time of the last predicted step, $\hat{y}_{i,t}$ and $y_{i,t}$ are the
predicted and ground truth positions of pedestrian $i$ at time step $t$,
respectively. Since we are interested in evaluating models that predict multiple
trajectories for each pedestrian, we keep just the minimum ADE for each agent $i$ (Best-of-N approach \cite{gupta2018social})
$\text{ADE}_{\text{min}}^i = \min_{k} \text{ADE}_{i}^{k}$,
where $\text{ADE}_{i}^{k}$ is the ADE of the $k$-th predicted trajectory of
pedestrian $i$, and finally take the average across the entire dataset.
Similarly, the Final Displacement Error (FDE) measures the L2 distance at the final predicted time step. Both are reported in meters.

\subsubsection{Environment Collisions-Free Likelihood (ECFL)}
\label{subsec:environment_collision_metric}

The ECFL metric \cite{sohn2021a2x} evaluates the ability of the model to predict collision-free trajectories. It is defined as the percentage of predicted trajectories that do not collide with obstacles in the scene (higher is better), indicating the model's ability to generate environment-compliant future paths.

The computation of ECFL is carried out by superimposing the predicted trajectories on a binary map representing the obstacles in the scene, and counting the number of predicted trajectories with at least one point over an obstacle. Then, we compute the percentage of trajectories that collide with the obstacles in the scene, by dividing the number of colliding trajectories by the total number of predicted trajectories (comprising all sampled paths for all pedestrians). This is the percentage of colliding samples, which is subtracted from $100\%$ to obtain the ECFL.

Formally, we denote the set of samples that collide with the obstacles in
the scene for a pedestrian $i$ as:
\begin{equation}
  C_i = \{k \in \{1, 2, \ldots, K\} \mid \text{collides}(\hat{Y}^k_i, I)\}
  \label{eq:metric_colliding_samples}
\end{equation}
where $\text{collides}(\hat{Y}^k_i, I)$ is the function
defined in (\ref{eq:metric_collides}).
Then, the Environment Collision-Free Likelihood for the whole dataset is computed as:
\begin{equation}
  \text{ECFL} = 100 - \frac{100}{N \times K} \sum_{i=1}^{N} |C_i|
  \label{eq:metric_environment_collision_complete}
\end{equation}
where $N \times K$ is the number of predicted trajectories in the whole
dataset ($K$ samples for all the $N$ pedestrians), and it reflects
the percentage of samples that don't collide with obstacles.

\subsection{\our Integration with other Models}
Our proposed approach requires the underlying model to take the map as input
and produce its embedding vector. SingularTrajectory
already supports image input, however, it uses a vector field to represent the
scene context, which is used when generating the prototype trajectories before
the diffusion refinement. We slightly modified the model to use the map embedding also in
the trajectory refinement process, in order to have a hidden representation
for the \mapnce module. EigenTrajectory-AgentFormer
already supports optional image conditioning, so we just needed to forward
the hidden representation to our module. EigenTrajectory-SGCN
does not directly support any map conditioning, so we slightly modified the
architecture to add the map embedding just before their TCN (time convolution network) \cite{bai2018empirical}
decoder, by simply summing the map embedding to the hidden representation, and
forwarding it to the \mapnce module.

For the underlying models that don't originally support map conditioning,
we used a pre-trained CNN, composed of four convolutional layers with ReLU activations, that embeds the input map into a 64-dimensional vector. The map patch surrounding each pedestrian is
rotated to align with the pedestrian's heading direction and offset to provide
a larger forward view of the scene. The CNN was trained in a self-supervised
manner using an autoencoder architecture 
(where the decoder module is another CNN with PixelShuffle as an upsampler \cite{shi2016real})
on a randomly generated set of map patches,
to learn to compress and reconstruct the patches, trying to
preserve the spatial structure of the scene.

The underlying models were trained using the best hyperparameters provided by the authors.
The three loss functions, namely the \mapnce loss, the Environmental Collision loss,
and the original loss of the underlying model, were combined through a weighted sum,
in such a way that the original loss and the environmental collision loss had
similar magnitudes, while the \mapnce loss had a magnitude of about a fourth of
the sum of the other two losses. These weights, and the other \our hyperparameters,
namely the temperature parameter $\tau = 0.5$, the number of negative samples $Z = 10$ per
pedestrian, the size of the contrastive embedding space for the query and key encoders set to $16$,
were chosen through hyperparameter tuning on the validation set.
During tuning, we observed that the model's collision avoidance capability was not overly sensitive to the exact value of these hyperparameters within a reasonable range, suggesting some flexibility in the parameters choice.

\subsection{Comparison with SOTA}

Table \ref{tab:env_col_results} and \ref{tab:ade_fde_results} report the results of the
comparisons. To ensure the robustness of the results, we report
the mean and standard deviation of the metrics calculated over $5$ runs.

\subsubsection{Evaluation of Collision Avoidance}
The collision avoidance capabilities, measured through the ECFL metric are reported in
Table \ref{tab:env_col_results}.
We compare the baseline models E-SGCN, E-AF, and ST with their enhanced versions incorporating \our.
The table indicates that adding \our to the base models significantly improves their
obstacle avoidance capabilities across all scenes. The average collision
percentage ($100\% - ECFL$) is effectively reduced by $43.01$\% for E-SGCN, $44.76$\% for E-AF, and $53.32$\% for ST.
Notably, ST+\our model consistently outperforms the other models, across all scenes,
yielding the highest average ECFL score ($96.06\pm0.17$) with a significant reduction in collisions.
Similar findings are observed in the qualitative results shown in Figure \ref{fig:qualitative_collisions},
where the paths generated by the enhanced models are more compliant with the environment.

While the ECFL score improves across all models, the inherent capability of the backbone architecture to handle fine-grained spatial constraints likely contributes to the final ECFL score. The iterative refinement process in diffusion (ST) may be particularly well-suited to integrating the spatial awareness learned via \our, leading to more effective collision avoidance compared to the single-pass decoding in Transformer- or GCN-based models.

\subsubsection{Evaluation of Trajectory Forecasting}
\begin{figure*}[th!]
  \centering
  %\framebox{\parbox{\textwidth}{Add here the pipeline of your module\\!\\!\\!\\!\\!\\!\\!\\!\\}}
  \includegraphics[width=0.92\textwidth]{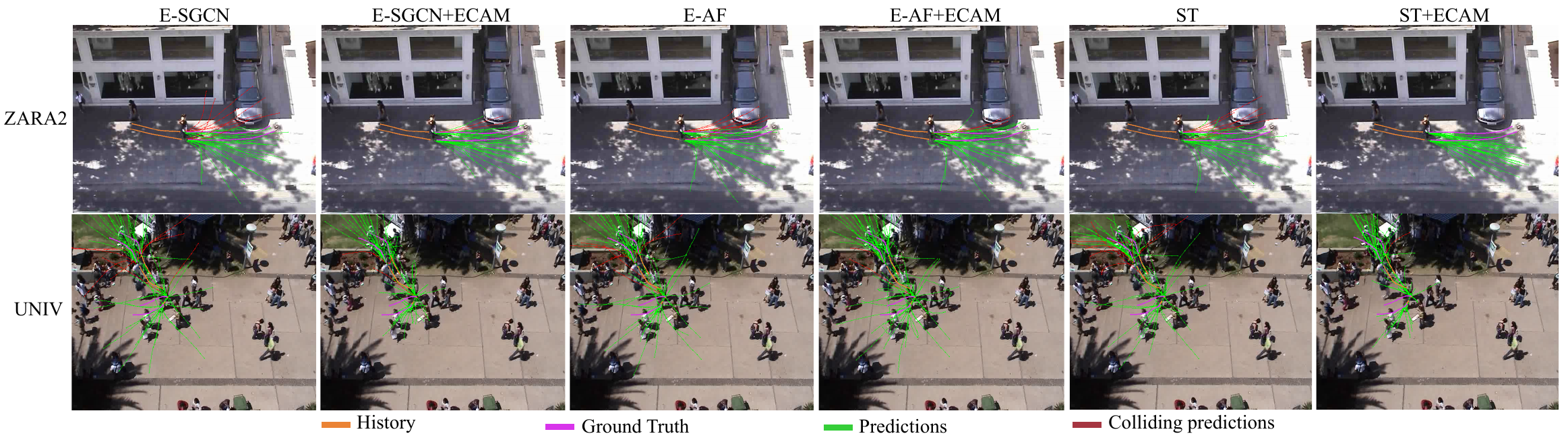}
  \caption{Qualitative results on ETH/UCY dataset when \our is integrated into E-SGCN, E-AF and ST and compared with original models. Models with \our show no or fewer collisions as compared to without \our. }
  \label{fig:qualitative_collisions}
\end{figure*}

Table \ref{tab:ade_fde_results} shows the $\text{ADE}_\text{min}/\text{FDE}_\text{min}$ results for the models with and without \our.
The inclusion of \our leads to a negligible decrease in performance in the $\text{ADE}_\text{min}/\text{FDE}_\text{min}$ metrics.
The average degradation for the $\text{ADE}_\text{min}$ metric is around 1-2 cm, while for the $\text{FDE}_\text{min}$ metric it is around 3-4 cm.
However, the models with \our still remain competitive with the original models.
The ST model without \our achieves the best performance in terms of $\text{ADE}_\text{min}/\text{FDE}_\text{min}$, while the E-SGCN+\our model
achieves the best performance across the models that employ \our.
The qualitative results shown in Figure. \ref{fig:qualitative_collisions} confirm that the best trajectories generated by the models with \our
are comparable to the original models, with the added benefit of collision avoidance.
In safety-critical applications like interactive robotics, avoiding collisions is more important than minimizing displacement error. A prediction that leads to a collision, even if close to the ground truth, is a critical failure. In contrast, a slightly less accurate but collision-free trajectory is far more valuable. The observed increase in ADE/FDE is typically within acceptable limits, given uncertainties in pedestrian behavior and sensor noise. In these contexts, the 40–50\% collision reduction achieved by \our offers greater practical value than marginal gains in ADE/FDE.
However, the acceptability of this trade-off can be application-dependent. For instance, in purely analytical tasks where trajectory prediction is used for behavior understanding without direct physical interaction, minimizing ADE/FDE might be the primary concern if collision-checking is handled separately or is less critical.

One possible reason for the performance decrease is the auxiliary Environmental Collision loss. This loss pulls colliding trajectories toward the only available ground truth, reducing sample diversity. As a result, errors increase in less common trajectories. Additionally, the \mapnce contrastive objective improves scene understanding but may not fully align with displacement error minimization. This misalignment could lead to minor optimization conflicts.

\begin{table*}[ht]
\caption{ECFL Results for different models across ETH/UCY dataset. \textbf{Bold} indicates overall best, \underline{underline} indicates best among same architecture, * means models are adapted to integrate map information for fair comparison.}
\label{tab:env_col_results}
\centering
\renewcommand{\arraystretch}{1.2}
\begin{tabular}{c|c|c|c|c|c|c|c}
\hline
Model & Backbone Architecture & \multicolumn{6}{c}{ECFL$\pm$STD $\uparrow$} \\ \cline{3-8}
          &   & ETH               & HOTEL             & UNIV              & ZARA1             & ZARA2             & AVG               \\ \hline
E-SGCN* \cite{bae2023eigentrajectory}\cite{shi2021sgcn}   & Graph Convolution       & 75.78$\pm$0.94   & 86.83$\pm$0.68   & 89.35$\pm$0.38   & 86.11$\pm$0.53   & 91.47$\pm$0.43    & 85.91$\pm$0.23 \\
E-SGCN+\our (Our)     &   & \underline{88.36$\pm$0.71}   & \underline{91.49$\pm$0.53}    & \underline{92.72$\pm$0.30}    & \underline{93.23$\pm$0.32}    & \underline{94.05$\pm$0.36}    & \underline{91.97$\pm$0.17}  \\ \hline
E-AF* \cite{bae2023eigentrajectory}\cite{yuan2021agentformer}  & Transformer  & 79.97$\pm$2.06   & 88.39$\pm$1.29   & 90.27$\pm$1.05    & 89.66$\pm$1.06   & 91.15$\pm$1.01    & 87.89$\pm$0.42 \\
E-AF+\our (Our) & & \underline{90.16$\pm$0.98}    & \underline{91.99$\pm$0.70}    & \underline{92.62$\pm$0.66}    & \underline{95.55$\pm$0.47}    & \underline{96.24$\pm$0.51}    & \underline{93.31$\pm$0.35}  \\ \hline
ST* \cite{bae2024singulartrajectory}    & Diffusion    & 83.47$\pm$0.63   & 92.43$\pm$0.51    & 90.46$\pm$0.37    & 95.17$\pm$0.21    & 96.25$\pm$0.19    & 91.56$\pm$0.11  \\
ST+\our  (Our)  & & \textbf{92.77}$\pm$\textbf{0.41}    & \textbf{94.80}$\pm$\textbf{0.37}    & \textbf{94.26}$\pm$\textbf{0.26}    & \textbf{99.32}$\pm$\textbf{0.14}    & \textbf{99.15}$\pm$\textbf{0.17}    & \textbf{96.06}$\pm$\textbf{0.17}  \\ \hline
\end{tabular}
\end{table*}

\begin{table*}[ht]
\caption{ADE/FDE results for different models across ETH/UCY dataset. \textbf{Bold} indicates overall best, \underline{underline} indicates best among same architecture, * means models are adapted to integrate map information for fair comparison.}
\label{tab:ade_fde_results}
\centering
\renewcommand{\arraystretch}{1.2}
\begin{tabular}{c|c|c|c|c|c|c|c}
\hline
Model & Backbone Architecture & \multicolumn{6}{c}{$\text{ADE}_\text{min}$/$\text{FDE}_\text{min}$ $\downarrow$} \\ \cline{3-8}
  &    & ETH               & HOTEL             & UNIV              & ZARA1             & ZARA2             & AVG               \\ \hline
E-SGCN*  \cite{bae2023eigentrajectory}\cite{shi2021sgcn}   & Graph Convolution            & \underline{0.37}/0.63             & \underline{0.13}/0.22             & \textbf{0.24}/\textbf{0.43} & 0.21/0.37                         & \underline{0.16}/0.28             & \underline{0.22}/0.39 \\
E-SGCN+\our (Our)                        &                              & \underline{0.37}/\underline{0.59} & \underline{0.13}/\underline{0.21} & 0.26/0.45                   & \underline{0.20}/\underline{0.35} & \underline{0.16}/\underline{0.26} & \underline{0.22}/\underline{0.37} \\ \hline
E-AF* \cite{bae2023eigentrajectory}\cite{yuan2021agentformer}    & Transformer      & \underline{0.37}/\underline{0.59} & \underline{0.14}/\underline{0.23} & \underline{0.25}/\underline{0.45} & \underline{0.21}/\underline{0.42} & \underline{0.17}/\underline{0.28} & \underline{0.23}/\underline{0.39} \\
E-AF+\our (Our)                  &                                      & 0.39/0.63 & 0.15/0.24 & \underline{0.25}/0.46 & \underline{0.21}/0.43 & 0.18/\underline{0.28} & 0.24/0.41 \\ \hline
ST* \cite{bae2024singulartrajectory}    & Diffusion                               & \textbf{0.36/0.44} & \textbf{0.13/0.21} & \underline{0.26}/\underline{0.45} & \textbf{0.19/0.33} & \textbf{0.15/0.26} & \textbf{0.22/0.34} \\
ST+\our (Our)                       &                                   & 0.37/0.51          & 0.15/0.24          & 0.27/0.47                         & 0.23/0.39          & 0.17/0.30          & 0.24/0.38 \\  \hline
\end{tabular}
\end{table*}

\begin{table*}[ht]
\caption{Ablation study of the individual components of the \our module.}
\label{tab:ablation}
\centering
\renewcommand{\arraystretch}{1.2}
\begin{tabular}{c|c|c|c|c|c|c}
\hline
Method & \multicolumn{6}{c}{$\text{ADE}_\text{min} \downarrow$ / $\text{FDE}_\text{min} \downarrow$ / ECFL $\uparrow$} \\ \cline{2-7}
      & ETH               & HOTEL             & UNIV              & ZARA1             & ZARA2             & AVG               \\ \hline
Baseline \cite{bae2024singulartrajectory} & 0.35/\textbf{0.42}/83.81 & \textbf{0.13}/\textbf{0.19}/92.62 &\textbf{0.25}/\textbf{0.44}/90.25 & \textbf{0.19}/\textbf{0.32}/95.31 & \textbf{0.15}/\textbf{0.25}/96.22 & \textbf{0.21}/\textbf{0.32}/91.64 \\
+MAP    & \textbf{0.34}/0.43/84.28    & 0.16/0.28/94.18    & 0.26/0.46/90.45    & 0.20/0.34/95.44    & 0.16/0.27/96.47    & 0.22/0.36/91.81  \\
+EnvColLoss    & 0.38/0.49/89.66    & 0.15/0.27/94.20    & 0.28/0.48/93.53    & 0.22/0.40/97.28    & 0.16/0.28/98.01    & 0.24/0.38/94.54  \\
+\mapnce    & 0.36/0.45/85.35    & 0.16/0.28/94.20    & 0.27/0.47/91.00    & 0.20/0.34/95.71    & 0.16/0.28/96.83    & 0.23/0.36/92.62  \\
+\our    & 0.38/0.51/\textbf{92.10}    & 0.14/0.23/\textbf{95.02}    & 0.29/0.50/\textbf{93.91}    & 0.22/0.39/\textbf{99.32}    & 0.18/0.30/\textbf{99.20}    & 0.24/0.38/\textbf{95.91}  \\ \hline
\end{tabular}
\end{table*}

\subsection{Ablation studies}

We provide ablation studies in Table \ref{tab:ablation} to showcase the improvements brought by each component of the proposed approach when integrated into the SingularTrajectory model (baseline).
For the purpose of discussion and analysis, we will refer to the percentage of colliding trajectories ($100\% - ECFL$) instead of the percentage of non-colliding ones ($ECFL$).
The baseline model uses the map information in an early step, but not on the actual trajectory refinement process. Although it achieves the lowest $\text{ADE}_\text{min}$/$\text{FDE}_\text{min}$, it has a high environment collision rate of $8.36$\%, underscoring its inability to avoid obstacles. When we condition the diffusion process of the model with the map embedding, the $\text{ADE}_\text{min}$/$\text{FDE}_\text{min}$ slightly increases, and the collision rate marginally decreases.
Incorporating environment loss (+EnvColLoss) significantly reduces the number of collisions to $5.46$\%, demonstrating better collision avoidance, though it will cost minimal increases in $\text{ADE}_\text{min}$ and $\text{FDE}_\text{min}$. Adding just \mapnce (without the EnvColLoss component), instead of EnvColLoss, also reduces the collision rate to $7.38$\%, balancing obstacle avoidance with moderate $\text{ADE}_\text{min}$/$\text{FDE}_\text{min}$ trade-offs.  The best results come from the integration of the full \our module, which reduces the percentage of collisions to $4.09$\% with a slight increase in $\text{ADE}_\text{min}$ to $0.24$ and $\text{FDE}_\text{min}$ to $0.38$. This shows that the combined use of \mapnce and EnvColLoss offers the most substantial improvement in collision avoidance with acceptable $\text{ADE}_\text{min}$/$\text{FDE}_\text{min}$ trade-offs of 3-6 cm.

The results suggest that the environmental collision loss is more effective in reducing collisions compared to using the \mapnce module alone. This difference in effectiveness can be explained by examining how each component influences the model's learning process. When \mapnce is used alone, it helps the model learn a good representation to distinguish between valid and obstacle-constrained trajectories. However, this learning doesn't effectively translate into better collision avoidance because the main task loss used by the underlying model (variety loss \cite{gupta2018social}) rarely provides relevant feedback for collision prevention, as the best trajectory is usually collision-free and far from obstacles. Variety loss, due to being implemented through the \emph{min} function, acts as a gradient router, only backpropagating gradients for the best trajectory.
Consequently, the model fails to meaningfully link the obstacle-aware representations learned via \mapnce to the collision avoidance task. Without explicit supervision from the environmental collision loss, which directly penalizes collisions across all trajectory samples, the model struggles to integrate collision avoidance into its prediction logic, even when equipped with \mapncenospace's discriminative capabilities.

\section{Limitations and Future Work}
The integration of \our into state-of-the-art methods significantly reduces collisions with the surrounding environment, enhancing the physical plausibility and safety of predicted trajectories. However, this enhancement introduces a minor increase in ADE/FDE, on the order of 1-2 cm for $\text{ADE}_\text{min}$ and 3-4 cm for $\text{FDE}_\text{min}$ in our experiments. This presents a trade-off between collision avoidance and raw displacement accuracy. Future research could focus on further refining this balance.
Future work could refine this balance through curriculum learning, where the model first emphasizes collision avoidance before focusing on precision, or by enhancing trajectory diversity using knowledge distillation. A teacher model could encourage diverse predictions, while \our ensures environmental compliance, helping capture multimodal human behaviors while maintaining reduced collision rates.

Currently, \our considers only static obstacles. A significant direction for future work is to extend its capabilities to handle dynamic obstacles, such as vehicles, which would require integrating temporal awareness of obstacle states into the contrastive learning framework.

\section{CONCLUSIONS}
In this paper, we introduced \our, a plug-n-play module designed to improve the environmental collision avoidance capabilities of trajectory forecasting models. By combining contrastive learning (\mapncenospace) with an explicit Environmental Collision loss, \our 
steers models toward generating trajectories that remain within navigable spaces, a key requirement for applications like autonomous driving and robotics. We demonstrated \our's effectiveness by integrating it into various state-of-the-art architectures. Evaluations on the ETH/UCY dataset, using the Environment Collision-Free Likelihood (ECFL) metric, quantitatively confirmed that \our substantially reduces collisions with the environment across diverse models.

\section{Acknowledgement}
This study was carried out within the project PRINPNRR22 “EasyWalk: Intelligent Social Walker for active living” and received funding from the European Union Next-GenerationEU - National Recovery and Resilience Plan (NRRP) – MISSION 4 COMPONENT 2, INVESTMENT 1.1 – CUP N. H53D23008220001. This manuscript reflects only the authors’ views and opinions, neither the European Union nor the European Commission can be considered responsible for them.

\addtolength{\textheight}{-0cm}   % This command serves to balance the column lengths

%%%%%%%%%%%%%%%%%%%%%%%%%%%%%%%%%%%%%%%%%%%%%%%%%%%%%%%%%%%%%%%%%%%%%%%%%%%%%%%%

\bibliographystyle{ieeetr}
\bibliography{main}

\begin{thebibliography}{10}

\bibitem{gulzar2021survey}
M.~Gulzar, Y.~Muhammad, and N.~Muhammad, ``A survey on motion prediction of pedestrians and vehicles for autonomous driving,'' {\em IEEE Access}, vol.~9, pp.~137957--137969, 2021.

\bibitem{socialrobot}
C.~Rösmann, M.~Oeljeklaus, F.~Hoffmann, and T.~Bertram, ``Online trajectory prediction and planning for social robot navigation,'' in {\em 2017 IEEE International Conference on Advanced Intelligent Mechatronics (AIM)}, pp.~1255--1260, 2017.

\bibitem{ano_pred}
W.~Liu, D.~L. W.~Luo, and S.~Gao, ``Future frame prediction for anomaly detection -- a new baseline,'' in {\em 2018 IEEE Conference on Computer Vision and Pattern Recognition (CVPR)}, 2018.

\bibitem{alahi2016social}
A.~Alahi, K.~Goel, V.~Ramanathan, A.~Robicquet, L.~Fei-Fei, and S.~Savarese, ``Social lstm: Human trajectory prediction in crowded spaces,'' in {\em Proceedings of the IEEE conference on computer vision and pattern recognition}, pp.~961--971, 2016.

\bibitem{sun2020recursive}
J.~Sun, Q.~Jiang, and C.~Lu, ``Recursive social behavior graph for trajectory prediction,'' in {\em Proceedings of the IEEE/CVF conference on computer vision and pattern recognition}, pp.~660--669, 2020.

\bibitem{zhang2019sr}
P.~Zhang, W.~Ouyang, P.~Zhang, J.~Xue, and N.~Zheng, ``Sr-lstm: State refinement for lstm towards pedestrian trajectory prediction,'' in {\em Proceedings of the IEEE/CVF Conference on Computer Vision and Pattern Recognition}, pp.~12085--12094, 2019.

\bibitem{transformer}
F.~Giuliari, I.~Hasan, M.~Cristani, and F.~Galasso, ``Transformer networks for trajectory forecasting,'' in {\em 2020 25th International Conference on Pattern Recognition (ICPR)}, pp.~10335--10342, 2021.

\bibitem{liu2021social}
Y.~Liu, Q.~Yan, and A.~Alahi, ``Social nce: Contrastive learning of socially-aware motion representations,'' in {\em Proceedings of the IEEE/CVF International Conference on Computer Vision}, pp.~15118--15129, 2021.

\bibitem{mangalam2020not}
K.~Mangalam, H.~Girase, S.~Agarwal, K.-H. Lee, E.~Adeli, J.~Malik, and A.~Gaidon, ``It is not the journey but the destination: Endpoint conditioned trajectory prediction,'' in {\em Computer Vision--ECCV 2020: 16th European Conference, Glasgow, UK, August 23--28, 2020, Proceedings, Part II 16}, pp.~759--776, Springer, 2020.

\bibitem{mangalam2021goals}
K.~Mangalam, Y.~An, H.~Girase, and J.~Malik, ``From goals, waypoints \& paths to long term human trajectory forecasting,'' in {\em Proceedings of the IEEE/CVF International Conference on Computer Vision}, pp.~15233--15242, 2021.

\bibitem{wang2022stepwise}
C.~Wang, Y.~Wang, M.~Xu, and D.~J. Crandall, ``Stepwise goal-driven networks for trajectory prediction,'' {\em IEEE Robotics and Automation Letters}, vol.~7, no.~2, pp.~2716--2723, 2022.

\bibitem{chen2023goal}
X.~Chen, F.~Luo, F.~Zhao, and Q.~Ye, ``Goal-guided and interaction-aware state refinement graph attention network for multi-agent trajectory prediction,'' {\em IEEE Robotics and Automation Letters}, vol.~9, no.~1, pp.~57--64, 2023.

\bibitem{gupta2018social}
A.~Gupta, J.~Johnson, L.~Fei-Fei, S.~Savarese, and A.~Alahi, ``Social gan: Socially acceptable trajectories with generative adversarial networks,'' in {\em Proceedings of the IEEE conference on computer vision and pattern recognition}, pp.~2255--2264, 2018.

\bibitem{dendorfer2021mg}
P.~Dendorfer, S.~Elflein, and L.~Leal-Taix{\'e}, ``Mg-gan: A multi-generator model preventing out-of-distribution samples in pedestrian trajectory prediction,'' in {\em Proceedings of the IEEE/CVF International Conference on Computer Vision}, pp.~13158--13167, 2021.

\bibitem{kosaraju2019social}
V.~Kosaraju, A.~Sadeghian, R.~Mart{\'\i}n-Mart{\'\i}n, I.~Reid, H.~Rezatofighi, and S.~Savarese, ``Social-bigat: Multimodal trajectory forecasting using bicycle-gan and graph attention networks,'' {\em Advances in neural information processing systems}, vol.~32, 2019.

\bibitem{bae2024singulartrajectory}
I.~Bae, Y.-J. Park, and H.-G. Jeon, ``Singulartrajectory: Universal trajectory predictor using diffusion model,'' in {\em Proceedings of the IEEE/CVF Conference on Computer Vision and Pattern Recognition}, pp.~17890--17901, 2024.

\bibitem{yue2022human}
J.~Yue, D.~Manocha, and H.~Wang, ``Human trajectory prediction via neural social physics,'' in {\em European conference on computer vision}, pp.~376--394, Springer, 2022.

\bibitem{yuan2021agentformer}
Y.~Yuan, X.~Weng, Y.~Ou, and K.~M. Kitani, ``Agentformer: Agent-aware transformers for socio-temporal multi-agent forecasting,'' in {\em Proceedings of the IEEE/CVF International Conference on Computer Vision}, pp.~9813--9823, 2021.

\bibitem{rempe2023trace}
D.~Rempe, Z.~Luo, X.~Bin~Peng, Y.~Yuan, K.~Kitani, K.~Kreis, S.~Fidler, and O.~Litany, ``Trace and pace: Controllable pedestrian animation via guided trajectory diffusion,'' in {\em Proceedings of the IEEE/CVF Conference on Computer Vision and Pattern Recognition}, pp.~13756--13766, 2023.

\bibitem{lee2022muse}
M.~Lee, S.~S. Sohn, S.~Moon, S.~Yoon, M.~Kapadia, and V.~Pavlovic, ``Muse-vae: Multi-scale vae for environment-aware long term trajectory prediction,'' in {\em Proceedings of the IEEE/CVF conference on computer vision and pattern recognition}, pp.~2221--2230, 2022.

\bibitem{liu2024trajdiffuse}
Q.~T. Liu, D.~Li, S.~S. Sohn, S.~Yoon, M.~Kapadia, and V.~Pavlovic, ``Trajdiffuse: A conditional diffusion model for environment-aware trajectory prediction,'' in {\em International Conference on Pattern Recognition}, pp.~382--397, Springer, 2024.

\bibitem{pellegrini2009you}
S.~Pellegrini, A.~Ess, K.~Schindler, and L.~Van~Gool, ``You'll never walk alone: Modeling social behavior for multi-target tracking,'' in {\em 2009 IEEE 12th international conference on computer vision}, pp.~261--268, IEEE, 2009.

\bibitem{lerner2007crowds}
A.~Lerner, Y.~Chrysanthou, and D.~Lischinski, ``Crowds by example,'' {\em Computer Graphics Forum}, vol.~26, 2007.

\bibitem{helbing1995social}
D.~Helbing and P.~Molnar, ``Social force model for pedestrian dynamics,'' {\em Physical review E}, vol.~51, no.~5, p.~4282, 1995.

\bibitem{van2008reciprocal}
J.~Van~den Berg, M.~Lin, and D.~Manocha, ``Reciprocal velocity obstacles for real-time multi-agent navigation,'' in {\em 2008 IEEE international conference on robotics and automation}, pp.~1928--1935, Ieee, 2008.

\bibitem{lee2017desire}
N.~Lee, W.~Choi, P.~Vernaza, C.~B. Choy, P.~H. Torr, and M.~Chandraker, ``Desire: Distant future prediction in dynamic scenes with interacting agents,'' in {\em Proceedings of the IEEE conference on computer vision and pattern recognition}, pp.~336--345, 2017.

\bibitem{sadeghian2019sophie}
A.~Sadeghian, V.~Kosaraju, A.~Sadeghian, N.~Hirose, H.~Rezatofighi, and S.~Savarese, ``Sophie: An attentive gan for predicting paths compliant to social and physical constraints,'' in {\em Proceedings of the IEEE/CVF conference on computer vision and pattern recognition}, pp.~1349--1358, 2019.

\bibitem{salzmann2020trajectron++}
T.~Salzmann, B.~Ivanovic, P.~Chakravarty, and M.~Pavone, ``Trajectron++: Dynamically-feasible trajectory forecasting with heterogeneous data,'' in {\em Computer Vision--ECCV 2020: 16th European Conference, Glasgow, UK, August 23--28, 2020, Proceedings, Part XVIII 16}, pp.~683--700, Springer, 2020.

\bibitem{vaswani2017attention}
A.~Vaswani, N.~Shazeer, N.~Parmar, J.~Uszkoreit, L.~Jones, A.~N. Gomez, {\L}.~Kaiser, and I.~Polosukhin, ``Attention is all you need,'' {\em Advances in neural information processing systems}, vol.~30, 2017.

\bibitem{franco2023under}
L.~Franco, L.~Placidi, F.~Giuliari, I.~Hasan, M.~Cristani, and F.~Galasso, ``Under the hood of transformer networks for trajectory forecasting,'' {\em Pattern Recognition}, vol.~138, p.~109372, 2023.

\bibitem{liu2021multimodal}
Y.~Liu, J.~Zhang, L.~Fang, Q.~Jiang, and B.~Zhou, ``Multimodal motion prediction with stacked transformers,'' in {\em Proceedings of the IEEE/CVF Conference on Computer Vision and Pattern Recognition}, pp.~7577--7586, 2021.

\bibitem{ho2020denoising}
J.~Ho, A.~Jain, and P.~Abbeel, ``Denoising diffusion probabilistic models,'' {\em Advances in neural information processing systems}, vol.~33, pp.~6840--6851, 2020.

\bibitem{nichol2021improved}
A.~Q. Nichol and P.~Dhariwal, ``Improved denoising diffusion probabilistic models,'' in {\em International conference on machine learning}, pp.~8162--8171, PMLR, 2021.

\bibitem{gu2022stochastic}
T.~Gu, G.~Chen, J.~Li, C.~Lin, Y.~Rao, J.~Zhou, and J.~Lu, ``Stochastic trajectory prediction via motion indeterminacy diffusion,'' in {\em Proceedings of the IEEE/CVF Conference on Computer Vision and Pattern Recognition}, pp.~17113--17122, 2022.

\bibitem{xue2018ss}
H.~Xue, D.~Q. Huynh, and M.~Reynolds, ``Ss-lstm: A hierarchical lstm model for pedestrian trajectory prediction,'' in {\em 2018 IEEE Winter Conference on Applications of Computer Vision (WACV)}, pp.~1186--1194, IEEE, 2018.

\bibitem{li2019conditional}
J.~Li, H.~Ma, and M.~Tomizuka, ``Conditional generative neural system for probabilistic trajectory prediction,'' in {\em 2019 IEEE/RSJ International Conference on Intelligent Robots and Systems (IROS)}, pp.~6150--6156, IEEE, 2019.

\bibitem{shukla2022from}
A.~Shukla, S.~Roy, Y.~Chawla, A.~Amalanshu, S.~Pandey, R.~Agrawal, A.~Uppal, V.~N, P.~Mondal, A.~Dasgupta, and D.~Chakravarty, ``From goals, waypoints \& paths to long term human trajectory forecasting,'' in {\em ML Reproducibility Challenge 2021 (Fall Edition)}, 2022.

\bibitem{bae2023eigentrajectory}
I.~Bae, J.~Oh, and H.-G. Jeon, ``Eigentrajectory: Low-rank descriptors for multi-modal trajectory forecasting,'' in {\em Proceedings of the IEEE/CVF International Conference on Computer Vision}, 2023.

\bibitem{shi2021sgcn}
L.~Shi, L.~Wang, C.~Long, S.~Zhou, M.~Zhou, Z.~Niu, and G.~Hua, ``Sgcn: Sparse graph convolution network for pedestrian trajectory prediction,'' in {\em Proceedings of the IEEE/CVF conference on computer vision and pattern recognition}, pp.~8994--9003, 2021.

\bibitem{sohn2021a2x}
S.~S. Sohn, M.~Lee, S.~Moon, G.~Qiao, M.~Usman, S.~Yoon, V.~Pavlovic, and M.~Kapadia, ``A2x: An agent and environment interaction benchmark for multimodal human trajectory prediction,'' in {\em Proceedings of the 14th ACM SIGGRAPH Conference on Motion, Interaction and Games}, pp.~1--9, 2021.

\bibitem{bai2018empirical}
S.~Bai, J.~Z. Kolter, and V.~Koltun, ``An empirical evaluation of generic convolutional and recurrent networks for sequence modeling,'' {\em arXiv preprint arXiv:1803.01271}, 2018.

\bibitem{shi2016real}
W.~Shi, J.~Caballero, F.~Husz{\'a}r, J.~Totz, A.~P. Aitken, R.~Bishop, D.~Rueckert, and Z.~Wang, ``Real-time single image and video super-resolution using an efficient sub-pixel convolutional neural network,'' in {\em Proceedings of the IEEE conference on computer vision and pattern recognition}, pp.~1874--1883, 2016.

\end{thebibliography}

\end{document}